\documentclass[11pt]{article}
\usepackage[final]{acl}
\usepackage{times}
\usepackage{latexsym}
\usepackage[T1]{fontenc}
\usepackage[utf8]{inputenc}
\usepackage{microtype}
\usepackage{inconsolata}
\usepackage{graphicx}
\usepackage{amsmath}
\usepackage{multirow}
\usepackage{booktabs}
\usepackage{listings}
\usepackage{appendix}
\makeatletter
\def\fps@table{t}
\makeatother

\title{Anonpsy: A Graph-Based Framework for Structure-Preserving De-identification of Psychiatric Narratives}

\author{
  Kyung Ho Lim, MD$^{1,2}$ \\
  \texttt{dlarudgh0627@gmail.com}
  \And
  Byung-Hoon Kim, MD, PhD$^{1,2,3,4}$\thanks{Corresponding author.} \\
  \texttt{egyptdj@yonsei.ac.kr}
  \AND
  \normalfont
  $^{1}$Department of Psychiatry, Yonsei University College of Medicine \\
  $^{2}$Institute of Behavioral Sciences in Medicine, Yonsei University College of Medicine \\
  $^{3}$Department of Biomedical Systems Informatics, Yonsei University College of Medicine \\
  $^{4}$Yonsei Institute for Digital Healthcare, Yonsei University
}

\begin{document}

\maketitle

\begin{abstract}
Psychiatric narratives encode patient identity not only through explicit identifiers but also through idiosyncratic life events embedded in clinical structure. Existing de-identification approaches, including PHI masking and LLM-based synthetic rewriting, operate at the text level and offer limited control over which semantic elements are preserved or altered. We introduce \textbf{Anonpsy}, a de-identification framework that reformulates the task as graph-guided semantic rewriting. Anonpsy (1) converts each narrative into a semantic graph encoding clinical entities, temporal anchors, and typed relations; (2) applies graph-constrained perturbations that modify identifying context while preserving clinical structure; and (3) regenerates text via graph-conditioned LLM generation. Evaluated on 90 clinician-authored psychiatric case narratives, Anonpsy preserves diagnostic fidelity while achieving consistently low re-identification risk under expert, semantic, and GPT-5-based evaluations. Compared with a strong LLM-only rewriting baseline, Anonpsy yields substantially lower semantic similarity and identifiability. These results demonstrate that explicit structural representations combined with constrained generation provide an effective approach to de-identification for psychiatric narratives.
\end{abstract}

\begin{figure}[t]
    \centering
    \includegraphics[width=\columnwidth]{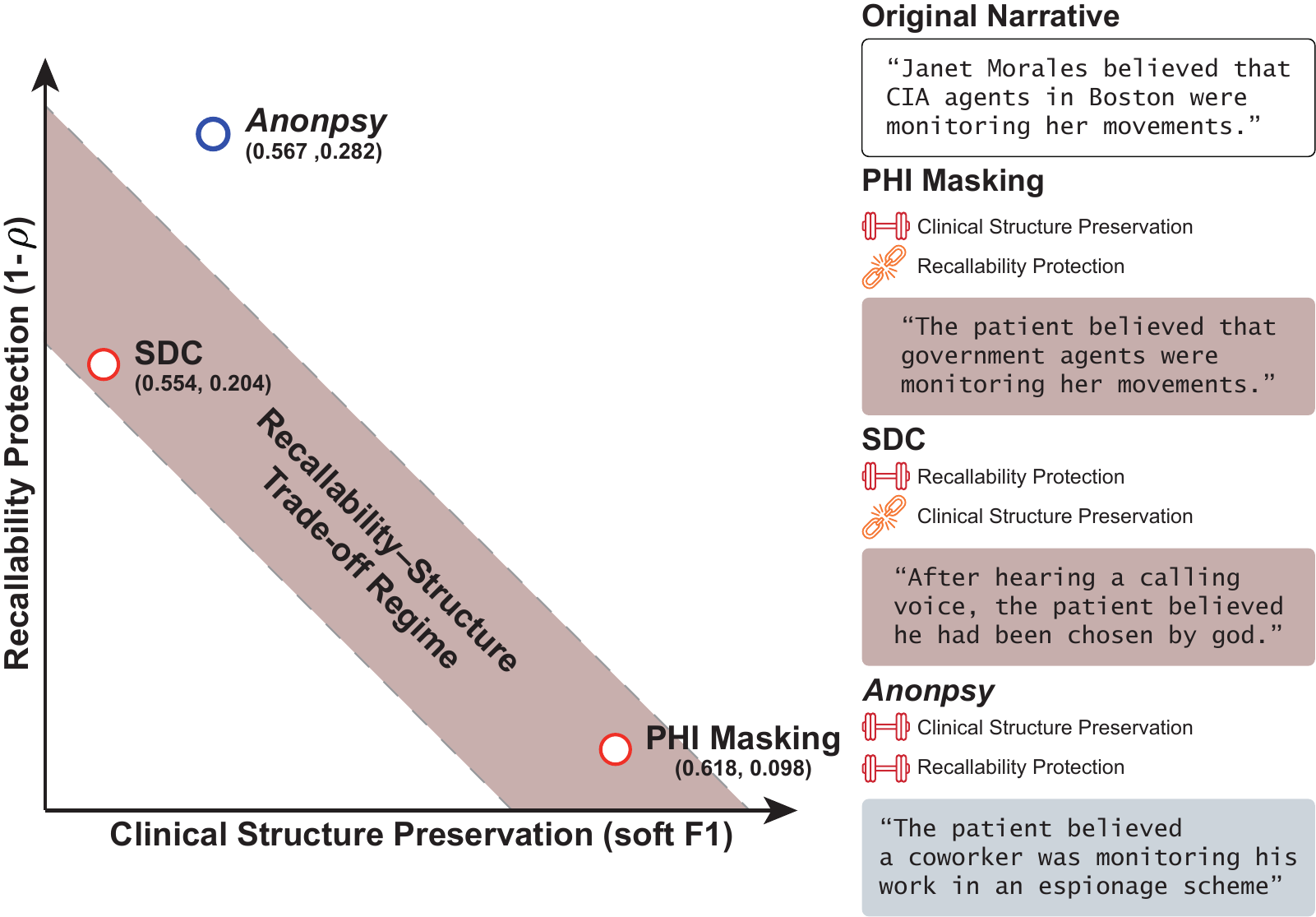}
    \caption{
Schematic illustration of the recallability--structure trade-off in psychiatric narrative de-identification. The horizontal axis represents preservation of clinical structure (approximated by diagnostic soft-F1), and the vertical axis represents protection against narrative recallability (inversely related to cosine similarity to the original narrative). Example excerpts illustrate how PHI masking preserves surface structure while retaining narrative identity, unconstrained SDC reduces recallability at the cost of semantic drift, and Anonpsy achieves a more balanced trade-off via structure-aware perturbation. (PHI: Protected Health Information; SDC: Synthetic Data Creation; $\rho$: Cosine similarity)
}
    \label{fig:schematic1}
\end{figure}

\section{Introduction}
Psychiatric narratives contain rich events, temporal and causal structure supporting downstream tasks such as diagnosis prediction and risk stratification~\citep{boland2021kaplan, DBLP:journals/corr/abs-2402-00179, stubbs2017identification}.

These characteristics make the de-identification of psychiatric narratives particularly challenging, as identifying signals are often embedded in narrative structure rather than limited to explicit identifiers~\citep{DBLP:journals/corr/abs-2402-00179, sarkar2025not}. As a result, existing approaches reflect an implicit recallability-structure trade-off (Figure~\ref{fig:schematic1}), where privacy risk manifests as the recallability of the original narrative. Token-level PHI masking preserves clinical structure but retains high semantic similarity, leaving substantial residual re-identification risk~\citep{stubbs2017identification, DBLP:journals/corr/abs-2402-00179, ford2025patient, kovavcevic2024identification}. In contrast, large language model-based synthetic data creation (LLM-based SDC) reduces recallability through broader rewriting but often distorts diagnostically essential structure~\citep{sarkar2025not}. We report a supporting trade-off premise check study in Appendix~\ref{sec:tradeoff}.

In psychiatric narratives, such distortions may alter diagnostic logic despite superficial plausibility; for example, transforming persecutory delusions into grandiose themes or introducing hallucination-like content absent from the source. More broadly, both paradigms treat clinical text as unstructured sequences, overlooking relational and temporal dependencies underlying psychiatric meaning \citep{tang2023does, altalla2025evaluating, fernandes2013development}.

Our approach occupies a different point in this trade-off space. We argue that de-identification for psychiatric narratives, which has high semantic identifiability, should be approached as a \emph{structure-preserving generation} problem. Here, \textit{structure} of a clinical narrative denotes an organized representation of clinical entities, including demographics, symptoms, treatments, and past history, enriched with temporal labels and interconnected through typed relations reflecting diagnostic, therapeutic, and causal dependencies. Our key insight is to represent each case as a \emph{semantic graph} with explicit temporal offsets and typed relations, enabling controlled perturbation of identifying content while preserving structure required for downstream clinical reasoning.

We introduce \textbf{Anonpsy}, a graph-guided de-identification framework that (1) converts the input psychiatric case text into a modifiable semantic graph of clinical entities; (2) performs controlled, graph-constrained perturbations of narrative elements; and (3) generates a coherent narrative via graph-conditioned text generation. By decoupling event structure from surface text, Anonpsy preserves clinically relevant structure while substantially reducing re-identification risk.

\paragraph{Contributions.} We make the following contributions:
\begin{itemize}
    \item We formulate psychiatric history text de-identification as a structured generation problem and introduce a dynamically editable semantic graph representation.
    \item We propose a graph-constrained perturbation and generation framework that selectively alters identifying narrative content while preserving clinically essential temporal, diagnostic, and causal structure.
    \item We provide comprehensive expert and GPT-5-based evaluations demonstrating improved privacy protection and structural fidelity compared to strong LLM-only rewriting baselines.
    \item We provide comprehensive expert, automated, and component-level ablation analyses demonstrating complementary roles of structural decomposition and perturbation ordering.
\end{itemize}

\section{Background and Related Work}

\subsection{Clinical Text De-identification}
Early work on clinical text de-identification treated the task as annotation and masking: rule-based pipelines, statistical models, and human annotators identified and suppressed PHI spans~\citep{dernoncourt2017identification}. PHI categories are typically defined by regulatory frameworks such as the Health Insurance Portability and Accountability Act and the General Data Protection Regulation~\citep{aghakasiri2025not}. Subsequent methods introduced neural architectures, including recurrent networks and transformer-based models trained on benchmark corpora such as i2b2 and CEGS N-GRID~\citep{johnson2020deidentification, uzuner2007evaluating}. These systems achieve high token-level accuracy but operate purely at the lexical level and do not model higher-order event structure~\citep{stubbs2017identification, aghakasiri2025not}.

\subsection{LLM-based De-identification and SDC}
The emergence of LLMs has enabled generative approaches that rewrite the entire note into a synthetic document rather than masking individual tokens~\citep{altalla2025evaluating, tang2023does}. These SDC methods can modify content beyond explicit identifiers, improving the privacy-utility trade-off compared with span masking~\citep{aghakasiri2025not}. However, they typically treat the note as an unstructured text sequence and expose limited control over which narrative elements are preserved or transformed, leading to unpredictable degradation or loss of clinically essential detail~\citep{sarkar2025not}. In psychiatric narratives, where clinical interpretation depends on longitudinal symptom development and distinguishing life events, such unconstrained rewriting can easily alter diagnostic logic even when the output appears superficially plausible.

\subsection{Structured Representations and Our Positioning}
Outside de-identification, biomedical natural language processing has explored structured and graph-based representations for tasks such as temporal relation extraction, entity linking, and knowledge-graph construction~\citep{sun2013evaluating, zhou2022clinical}. These systems typically build entity-relation or ontology graphs that are useful for retrieval and inference, but are not designed for narrative generation. Intermediate structures are rarely exposed to end users, and there is little support for manual inspection or selective modification of the underlying graph.

No prior work has introduced a de-identification framework for psychiatric narratives that uses a temporal event graph with typed relations as an intermediate representation for controlled generation. Existing de-identification and synthetic data approaches either (i) operate at the token level and ignore higher-level clinical structure, or (ii) rely on direct text-to-text rewriting without a controllable intermediate representation~\citep{sarkar2025not, moharasan2019extraction}. Psychiatric case histories are a particularly challenging setting: their clinical significance is deeply intertwined with personal life circumstances, longitudinal trajectories, and causal links among symptoms, treatments, and prior diagnoses~\citep{boland2021kaplan}. Our work addresses this gap by treating de-identification as a structure-preserving generation problem: we convert the narrative into an explicit temporal event graph, apply clinically constrained perturbations at the graph level, and then regenerate the narrative from this edited structure.

\section{Method}
\label{sec:method}

\subsection{Problem Definition}
\label{sec:problem}

We formulate de-identification as a \emph{structure-preserving generation} task. Given a narrative $X$, the goal is to produce a transformed narrative $\hat{X}$ that (i) minimizes re-identification risk and (ii) preserves clinical structure for downstream reasoning.
 
We convert each narrative into a semantic graph
\begin{align}
    G = \mathcal{E}(X) = (V, E),    
\end{align}

where $\mathcal{E}$ is an LLM-assisted, schema-constrained conversion operator that converts the raw narrative $X$ into a structured representation. $V$ contains clinical entities such as symptoms, treatments, and diagnoses, while $E$ contains typed relations.

We then apply a schema-constrained perturbation operator $\mathcal{P}$, which modifies contextual attributes of $G$ while preserving temporal, causal, and diagnostic structure:
\begin{align}
    \tilde{G} = \mathcal{P}(G).    
\end{align}

Finally a graph-conditioned text generation operator $\mathcal{D}$ then produces a de-identified narrative $\hat{X}$ from $\tilde{G}$:

\begin{align}
    \hat{X} = \mathcal{D}(\tilde{G}).
\end{align}

All three operators $(\mathcal{E}, \mathcal{P}, \mathcal{D})$ are implemented as hybrid systems combining deterministic rules with LLM-based, schema-constrained prompting; full implementation details are provided in Appendix~\ref{sec:method_detail}.

The objective is twofold:
\begin{enumerate}
    \item \textbf{Preserve clinical structure}: maintain the temporal alignment of events, and the diagnostic, therapeutic, and causal relations encoded in its structure so that $\hat{X}$ remains diagnostically faithful to $X$;
    \item \textbf{Reduce identifiability}: alter contextually identifiable life events and psychosocial details while ensuring that $\hat{X}$ cannot be used to infer the original patient.
\end{enumerate}

\begin{figure}[!ht]
    \centering
    \includegraphics[width=0.95\columnwidth]{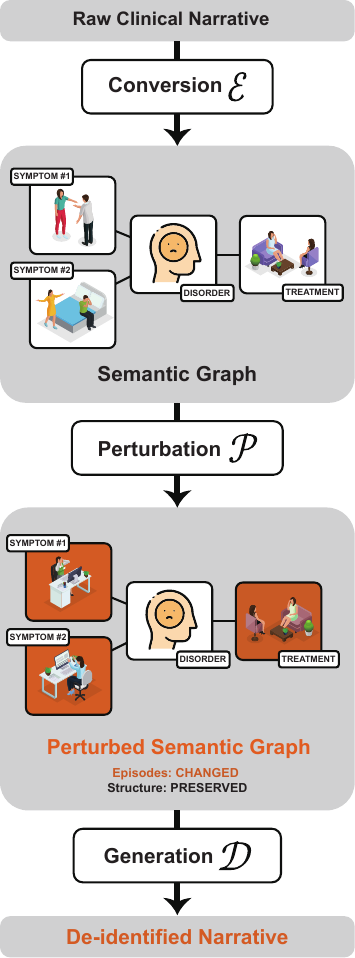}
    \caption{
Overview of the Anonpsy pipeline for de-identification of psychiatric narratives via structure-preserving generation. The conversion operator transforms a free-text psychiatric narrative into a schema-constrained semantic graph; the perturbation operator selectively edits identifiable contextual elements while preserving temporal structure and typed relations; and the generation operator produces the final de-identified narrative.
    }
    \label{fig:schematic2}
\end{figure}

This formulation differs from PHI masking and prior SDC approaches by introducing an explicit intermediate representation and enforcing structural constraints during generation. Figure~\ref{fig:schematic2} depicts the high-level workflow of the Anonpsy pipeline.

\subsection{LLM backbone}
All LLM-assisted components of the Anonpsy pipeline, including semantic graph conversion ($\mathcal{E}$), graph-based perturbation ($\mathcal{P}$), and graph-conditioned text generation ($\mathcal{D}$), use a single locally deployed LLM, \texttt{gpt-oss:120b} ~\citep{agarwal2025gpt}. Local deployment is often required in privacy-sensitive clinical environments where cloud-based LLM APIs are restricted by policy ~\citep{liu2025generating}. Each operator is executed within a distinct deterministic control flow that enforces schema and temporal constraints; the LLM is used only for bounded semantic inference. No proprietary or closed API systems were used.

The model is accessed via a ChatOllama backend. Lower decoding temperatures are used for schema-constrained extraction and narrative realization to prioritize stability and clinical fidelity, whereas a higher temperature is used for perturbation to encourage semantic diversity when rewriting identifying narrative content. Decoding hyperparameters are summarized in Appendix~\ref{sec:hyperparams} (Table~\ref{tab:hyperparams}). This model was run on a local machine equipped with four NVIDIA RTX A6000 GPUs (each with 48 GB VRAM). 

\subsection{Dataset}
We evaluate Anonpsy on \textit{DSM-5-TR Clinical Cases}, a publicly available collection of clinician-authored psychiatric narratives~\citep{barnhill2023dsm}. The corpus is published by the American Psychiatric Association and is subject to copyright. We use the dataset solely for analysis and evaluation within this study and do not redistribute the original narratives. All reported examples are presented in synthetic form. Our use falls under scholarly analysis of publicly available educational material and does not provide access to copyrighted content. The cases are well-suited because they contain specific psychosocial detail, exhibit nonlinear temporal structure, and span diverse diagnoses with frequent comorbidity.

Of the 92 available cases, we exclude two multi-person narratives, a couple-therapy case and a shared-trauma vignette, since clinical documentation is typically organized at the single-patient record level, and our framework assumes one patient per record.

\subsection{Semantic Graph Conversion}
We convert each narrative $X$ into a semantic graph $G$, in which clinical entities are represented as nodes and inter-entity relations as typed edges via $\mathcal{E}$. The conversion operator $\mathcal{E}$ is a schema-constrained, LLM-assisted module that maps a free-text psychiatric narrative into a structured semantic graph by identifying clinical entities, temporal attributes, and typed relations. The resulting graph serves as the standardized intermediate representation for perturbation and graph-conditioned text generation. 
% A synthetic YAML example illustrating the resulting graph structure appears in the Appendix~\ref{serial}.

\paragraph{Conversion pipeline.}
The semantic graph is constructed through a staged, schema-constrained conversion pipeline that combines LLM-assisted extraction with deterministic normalization and validation. The pipeline proceeds in four stages: (1) schema-guided entity extraction; (2) temporal episode construction and normalization; (3) deterministic temporal reconciliation; and (4) typed relation construction under schema constraints. In practice, the initial extraction stage uses schema-constrained LLM prompts to populate the predefined node types and attributes of the semantic graph (e.g., \textit{SymptomNode}, \textit{DiagnosisNode}, \textit{TreatmentNode}). The resulting structured outputs are then passed through deterministic normalization steps that convert temporal references into relative offsets anchored to the \textit{VisitEvent}, enforce relation-type constraints, and reconcile node durations with associated temporal spans before constructing the final graph $G$. This staged design separates semantic inference from deterministic temporal normalization, ontology-constrained relation validation, and node–duration reconciliation, enabling intermediate validation, limiting error propagation from LLM extraction steps and enforcing structural consistency before downstream perturbation and generation.
% Full conversion schemas and implementation details are provided in
% Appendix~\ref{sec:conversion}.

\paragraph{Semantic graph definition.}
\subparagraph{Node types.}
We instantiate nodes for diagnoses, symptoms, treatments, past history, and a structured \texttt{VisitEvent}. The \texttt{VisitEvent} node represents the most recent clinical encounter and anchors all temporal offsets; it encodes setting, arrival mode, and safety-relevant context, and links to currently active symptoms via \texttt{PRESENTS\_WITH} edges. Symptom episodes are decomposed using a modular Situation-Thought-Emotion-Behavior (STEB) schema, derived from the cognitive model of psychopathology, which isolates manipulable contextual elements while preserving clinically meaningful structure ~\citep{beck1979cognitive}.

\subparagraph{Temporal representation.}
Each event is anchored to a pool of duration objects capturing onset offset and span relative to an index encounter at day~0. Offsets before and after day~0 are encoded as signed integers, enabling ordering without specifying calendar dates. Nodes may reference one or more durations, allowing reasoning about temporal sequence, overlap, and whether a symptom is current or historical.

\subparagraph{Relation types.}
Edges encode clinically interpretable dependencies, including \texttt{MANIFESTS\_AS} (symptom to diagnosis), \texttt{TREATMENT\_OF} (treatment to symptom or diagnosis), \texttt{PRESENTS\_WITH} (visit event to current symptoms), and \texttt{INDUCES} (etiologic entity to diagnosis). Permissible source-target type pairs are fixed in advance to prevent unsupported or spurious edges.

\subsection{Graph-based Perturbation}

We apply a perturbation operator $\mathcal{P}$ to the semantic graph $G$ to produce a modified graph $\tilde{G}$ that reduces re-identification risk by altering identifying narrative content while preserving clinically meaningful structure. Perturbation is restricted to attribute and context fields and does not modify the temporal backbone or the typed inter-entity relations. In implementation, attribute perturbations are applied through deterministic transformation rules for structured fields (e.g., age offsets and laboratory value substitution), while contextual narrative fields within the STEB schema are modified through constrained LLM rewriting conditioned on the corresponding graph nodes. This ensures perturbation operates on graph-validated structures rather than raw text. As a result, temporal ordering, event overlap, and clinically constrained dependencies encoded in $G$ are preserved by construction in $\tilde{G}$.

Specifically, $\mathcal{P}$ perturbs (i) demographic attributes under clinically informed feasibility constraints; for example, age perturbations must remain consistent with onset thresholds and duration requirements of DSM-5-TR diagnoses, and sex perturbations respect sex-specific diagnostic labels. (ii) Symptom-associated contextual narratives represented in the STEB schema while preserving diagnostic meaning and temporal anchors, and (iii) numeric test-result values through range-preserving substitution that maintains clinical interpretation. 
% Full perturbation rules and implementation details are provided in Appendix~\ref{sec:perturb}.

Perturbation policies are constrained by clinically informed rules rather than learned optimization. Psychiatric diagnosis in DSM-5-TR is defined through structured criteria, such as symptom counts, duration thresholds, age-of-onset constraints, and causal exclusions, where small violations can render a case diagnostically implausible. For this reason, Anonpsy adopts deterministic, diagnosis-informed perturbation constraints to preserve temporal alignment and causal structure during rewriting.

\subsection{Graph-conditioned Text Generation}

A graph-conditioned text generation operator $\mathcal{D}$ produces a de-identified narrative $\hat{X}$ conditioned on the perturbed graph $\tilde{G}$. Events are sorted by temporal anchor and associated diagnoses to construct a deterministic timeline. Within each temporal block, symptoms are grouped by their linked diagnosis, and treatments targeting the same diagnosis and overlapping in duration are narrated within the same block. The perturbed graph $\tilde{G}$ is serialized into a structured prompt containing node attributes, temporal offsets, and typed relations. This representation conditions the LLM generation process so that narrative realization follows the temporal ordering and clinical relations encoded in $\tilde{G}$ without introducing new entities or altering graph structure. This grouping procedure derives directly from normalized temporal offsets and typed graph relations rather than unconstrained narrative segmentation. The model is prompted to preserve causal dependencies, avoid hallucinated clinical details, and maintain stylistic coherence. Because the generation process is conditioned on $\tilde{G}$, the resulting narrative $\hat{X}$ preserves the clinical structure of the original while replacing identifying contextual content. 
% Full details appear in the Appendix~\ref{sec:textgen}.

\section{Experiments}

We evaluate Anonpsy in terms of the two objectives introduced in Section~\ref{sec:problem}: (1) clinical structure preservation and  
(2) de-identification robustness. Because psychiatric diagnosis depends on temporal course, symptom configuration, and causal relations, distortions to diagnostically essential structure should manifest as shifts in downstream diagnostic judgments; we therefore use diagnosis preservation as a functional proxy for clinical structure preservation.
 
In addition to expert and GPT-5 evaluations, we quantify a deterministic proxy for \emph{narrative recallability} using embedding-based semantic cosine similarity between each original narrative and its de-identified variant. Methods that leave idiosyncratic life events and narrative content largely intact remain closer in embedding space due to higher semantic overlap, whereas stronger contextual perturbation yields lower similarity. We use this measure to provide a deterministic complement to LLM-based judgments.
% (Appendix~\ref{sec:cosine}).

We compare three narrative variants:  
(a) the original narrative from DSM-5-TR Clinical Cases (\emph{Original}),  
(b) the de-identified narrative produced by Anonpsy (\emph{Anonpsy}), and  
(c) a strong prompt-engineered rewrite generated by the same LLM without access to the graph structure (\emph{LLM-only}).  

The LLM-only condition serves as a strong text-only baseline implemented with the same backbone model and decoding configuration as Anonpsy. The baseline is carefully prompt-engineered to preserve diagnoses, temporal ordering, causal relations, and symptom trajectories while explicitly modifying demographics, locations, interpersonal details, and distinctive life events. It also includes a self-critique refinement stage that removes residual identifiable details and repairs structural inconsistencies. This design intentionally isolates a single factor: the presence or absence of explicit graph-based structural constraints. Full prompt details are provided in Appendix~\ref{sec:llmonly}.
% Full details are given in the Appendix~\ref{sec:llmonly}.

To obtain both expert-grounded and large-sample evidence, we conduct:
(i) a structured expert evaluation with five board-certified psychiatrists on 20 cases, and  
(ii) a GPT-5-based automatic evaluation on all 90 cases.  
Detailed protocols, model specifications, and statistical analyses appear in Appendix~\ref{sec:exp-details}.
    
\subsection{Expert Evaluation}
\label{sec:human-eval}

Five board-certified psychiatrists were recruited via social media; board certification was verified at enrollment, and no identifying documentation was retained beyond confirmation of eligibility. Participants provided informed consent and received a nominal honorarium for expert consultation. For each experiment, raters were presented with the narratives in a randomized, blinded order.

\subsubsection{Clinical Structure Preservation}

Participants were informed that the goal was to assess the clinical validity of synthetic patient data; the existence of parallel versions per case was not disclosed. For a randomly sampled 10 cases, each rater viewed all three variants and answered:

\begin{itemize}
    \item[\textbf{Q1}] "Please write the diagnosis that you consider most appropriate for this case."
    \item[\textbf{Q2}] "Do you think \{\texttt{ground truth diagnosis}\} is consistent with the clinical content of the case?" (yes/no)
\end{itemize}

Q1 responses were evaluated using a soft-F1 metric based on semantic matching. Differences across narrative types were analyzed using a linear mixed-effects model with random intercepts for case and rater, treating \emph{Anonpsy} as the reference condition. Q2 responses were analyzed using a Bayesian logistic mixed-effects model to handle strong class imbalance (most responses were "yes"). 
% Full model specifications are given in Appendix~\ref{sec:human_clinical}.

\paragraph{Results.}
Fixed-effect estimates for LLM-only versus Anonpsy ($\beta = 0.024$, $p = 0.699$) and Original versus Anonpsy ($\beta = 0.005$, $p = 0.932$) were near zero and non-significant, and raw group means were similar (Anonpsy: 0.719; LLM-only: 0.743; Original: 0.725) (Table~\ref{tab:q1_mixed_model}). Complementary Wilcoxon tests on per-case and per-input type averages likewise found no significant differences.  

\begin{table}[t]
\centering
\small
\begin{tabular}{lccc}
\toprule
\textbf{Effect} & \textbf{$\beta$} & \textbf{SE} & \textbf{$p$} \\
\midrule
Intercept (Anonpsy)       & 0.719 & 0.053 & <.001 \\
LLM-only vs.\ Anonpsy     & 0.024 & 0.062 & 0.699 \\
Original vs.\ Anonpsy     & 0.005 & 0.062 & 0.932 \\
\bottomrule
\end{tabular}
\caption{Mixed-effects model for Q1 soft-F1 scores. Fixed-effect estimates compare narrative types (Anonpsy as reference).}
\label{tab:q1_mixed_model}
\end{table}

For Q2, all narrative types were rated as highly consistent with their diagnoses: the mean probability of a "yes" response was 0.900 for \emph{Anonpsy} and 0.960 for both \emph{Original} and \emph{LLM-only}. Posterior 95\% credible intervals for all narrative-type contrasts included zero (Table~\ref{tab:q2_bayes}).

\begin{table}[t]
\centering
\small
\begin{tabular}{lcc}
\toprule
\textbf{Effect} & \textbf{Posterior Mean} & \textbf{Posterior SD} \\
\midrule
Intercept (Anonpsy)       & 2.444 & 0.345 \\
LLM-only vs.\ Anonpsy     & 1.097 & 0.669 \\
Original vs.\ Anonpsy     & 1.097 & 0.669 \\
\bottomrule
\end{tabular}
\caption{Bayesian logistic mixed-effects model for Q2 (narrative--diagnosis congruence).}
\label{tab:q2_bayes}
\end{table}

\paragraph{Interpretation.}
Across both diagnostic tasks, no statistically significant differences were observed between \emph{Anonpsy}, \emph{Original}, and \emph{LLM-only}. Fixed-effect contrasts were small and non-significant, and posterior uncertainty overlapped across conditions, indicating that \textbf{graph-based de-identification preserves diagnostic usability} under expert evaluation.

\subsubsection{Re-identification Robustness}

In this experiment, participants were informed that the goal was to assess the privacy properties of synthetic patient data. Detailed rubrics and model specifications are given in Appendix~\ref{sec:rubric}. In an experiment with 10 sampled cases without replacement, raters first read the original narrative, then a randomized A/B pair consisting of the Anonpsy and LLM-only versions of the same case. For each pair, raters answered 1) Which version (A or B) feels more similar to the original? 2) What is the re-identification risk of version A (1--5)? 3) What is the re-identification risk of version B (1--5)? In addition, raters provided brief free-text explanations justifying their similarity choice and risk scores for each version. These qualitative comments were collected to enable qualitative analysis of risk judgments and sources of perceived recallability. 

The 1--5 re-identification rubric (Appendix~\ref{sec:rubric}; Table~\ref{tab:rubric}) distinguishes generic, non-recallable narratives (scores 1--2) from narratives that a knowledgeable insider could plausibly link back to a specific individual (scores $\geq 3$). Paired risk scores were compared using a Wilcoxon signed-rank test; we additionally examined scores separately for the versions chosen as more similar and less similar to the original.

\paragraph{Results.}
Across 50 rater-case evaluations, Anonpsy outputs showed consistently low absolute re-identification risk. The vast majority of Anonpsy ratings fell in the "no or minimal risk" range (scores 1--2), with a median of~1. Importantly, \textbf{no Anonpsy narrative received a score above~3}, and \textbf{no case showed majority agreement among raters that the Anonpsy output was at moderate risk} (score $\geq 3$). Only one case received a score of ~3 by two raters.

In contrast, LLM-only rewrites exhibited a substantially heavier upper tail of risk scores. The mean paired risk score was 2.800 for LLM-only versus 1.540 for Anonpsy and median risk score was 3.000 and 1.000, respectively. A Wilcoxon signed-rank test confirmed the difference was highly significant ($W = 84.5$, $p < 0.001$), showing that LLM-only outputs were systematically rated as higher risk when evaluated on the same underlying case by the same psychiatrist (Table~\ref{tab:human_risk}).

\begin{table}[t]
\centering
\small
\begin{tabular}{lcccc}
\toprule
\textbf{Method} & \textbf{Mean} & \textbf{Median} & \textbf{SD} &
\textbf{Wilcoxon $p$} \\
\midrule
LLM-only & 2.800 & 3.000 & 1.195 & \multirow{2}{*}{$< 0.001$} \\
Anonpsy  & 1.540 & 1.000 & 0.646 & \\
\bottomrule
\end{tabular}
\caption{Paired re-identification risk scores (1--5) across 50 rater--case
evaluations. Wilcoxon signed-rank test compares LLM-only vs.\ Anonpsy.}
\label{tab:human_risk}
\end{table}

\begin{figure}[t]
    \centering
    \includegraphics[width=0.45\textwidth]{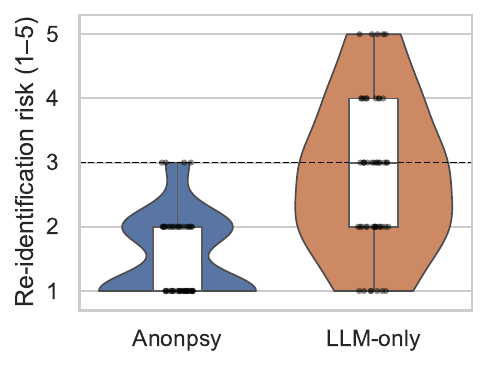}
    \caption{
        Violin plot of human-rated re-identification risk scores (1--5) for paired Anonpsy and LLM-only narratives. The horizontal line at score~3 indicates the predefined "at-risk" threshold.}
    \label{fig:risk_violin}
\end{figure}

When a version was chosen as more similar to the original, its mean risk score was 3.122 for LLM-only ($n=41$) versus 2.111 for Anonpsy ($n=9$), demonstrating a significant statistical difference (Mann--Whitney $U = 280.0$, $p = 0.012$). For non-chosen versions, mean risks were low and nearly indistinguishable (LLM-only: 1.333; Anonpsy: 1.415; $p = 0.731$), indicating that risk differences emerge specifically in the narratives that evoke the original story. Consistent with this, LLM-only was selected as more similar to the original in 41 of 50 trials (82\%; binomial $p < 0.001$).

\paragraph{Qualitative analysis of moderate-risk ratings.}
Although a small number of Anonpsy outputs received a moderate risk score of 3, qualitative review showed that these ratings did \emph{not} reflect narrative resemblance. In the only case where two raters gave a score of~3, both explained that their judgment was driven by the presence of a rare diagnosis (Lyme disease), noting that clinicians often refer to such cases by the rare diagnosis itself, making the case feel recognizable regardless of narrative content. Raters emphasized that the rewritten text did \emph{not} preserve distinctive events, quotations, or locations.

In the remaining instances, raters who judged Anonpsy as more similar than LLM-only attributed this to preserved \emph{clinical} similarity: shared symptom themes, developmental patterns, or broadly similar demographic information, which the pipeline intentionally maintains. Comments noted that the narratives were "not specific enough to identify an individual," that similarity arose from "general clinical patterns rather than details," and that "despite coherence, it does not feel like the same person." No rater ever identified an unchanged concrete anchor such as exact date, location, or unique event in an Anonpsy output. For a concrete illustrative example, including a case-level comparison of the Original, LLM-only, and Anonpsy outputs and representative evaluator comments, see Appendix~\ref{app:qualitative-example}.

\paragraph{Interpretation.}
Taken together, these findings show that Anonpsy achieves \textbf{very low face-value re-identification risk}. Ratings were strongly concentrated at 1--2, no case drew majority agreement for moderate risk, and no output exceeded a score of~3. Moderate ratings were driven by rare clinical features of the original case rather than preservation of narrative content. In contrast, LLM-only rewrites were frequently judged as similar to the original story and received substantially higher re-identification scores. Thus, Anonpsy successfully suppresses recognizable narrative identity while retaining the clinical information necessary for downstream reasoning.

\subsection{Automated GPT-5 Evaluation}
\label{sec:gpt-summary}
To verify that the human-evaluated patterns generalize across the full corpus, we conducted a parallel GPT-5-based evaluation on all 90 DSM-5-TR cases. GPT-5 was used both as a diagnostic reasoner and as a proxy for a knowledgeable insider. In both tasks, diagnosis prediction, and diagnosis acceptability, GPT-5 showed no significant differences between Anonpsy, the original narratives, and the LLM-only baseline after DSM-aware canonicalization, mirroring the human findings that Anonpsy preserves diagnostic structure. For de-identification robustness, embedding-based semantic similarity and a GPT-5 insider simulation both favored Anonpsy: its narratives were significantly more semantically distant from the originals and received lower re-identification risk scores than LLM-only rewrites (Tables~\ref{tab:cosine-sim} and~\ref{tab:reid-risk}). Full canonicalization rules, statistical tests, and detailed results are provided in Appendix~\ref{sec:gpt:diag:pred},~\ref{sec:gpt:diag:acc} 

\subsection{Component-Level Ablation Analysis}
\label{sec:ablation}

To quantify the contribution of individual design choices, we conducted targeted ablations isolating (i) STEB decomposition and (ii) retrograde perturbation ordering. All ablations were evaluated under the same GPT-5-based protocol as the full Anonpsy pipeline (Appendix~\ref{sec:gpt:diag:pred},~\ref{sec:gpt:diag:acc},~\ref{sec:cosine},~\ref{sec:gpt:reid}); full experimental details are provided in Appendix~\ref{sec:ablation_details}.

\paragraph{STEB Decomposition Ablation (NoSTEB).}
We replaced the STEB schema with a single free-text episode sentence per symptom node and applied perturbation directly to that sentence while preserving all other components. Diagnostic predictability (soft-F1) and diagnostic acceptability showed no significant differences from the full model (Wilcoxon $p=0.981$; McNemar $p=0.677$). However, semantic cosine similarity increased significantly (0.718 $\rightarrow$ 0.746, $p<0.001$), and GPT-5 re-identification risk scores were significantly higher ($p<0.001$). This indicates that STEB reduces narrative recallability while preserving clinical structure.

\paragraph{Retrograde Perturbation Ablation (NoRetro).}
We removed retrograde ordering and perturbed symptom contexts in chronological order while retaining the STEB schema. This significantly reduced diagnostic predictability (soft-F1: 0.575 $\rightarrow$ 0.525, $p=0.048$) and diagnostic acceptability (0.765 $\rightarrow$ 0.644, $p=0.032$), with no significant change in cosine similarity ($p=0.305$) and re-identification risk score ($p=0.748$). These results indicate that retrograde ordering primarily supports structural coherence and downstream diagnostic preservation.

\paragraph{Interpretation.}
The two components play complementary roles: STEB decomposition enhances privacy robustness, whereas retrograde perturbation preserves clinical structure. Full tables and statistical details are provided in Appendix~\ref{sec:ablation_details}.

\subsection{Error Analysis}
\label{sec:error_analysis}

We conducted a qualitative error analysis on representative cases to identify failure modes in graph construction and regeneration. The primary source of error arises from conservative entity inclusion during the conversion stage, where clinically relevant but diagnostically ambiguous concepts may be retained. For example, symptom-like descriptors (e.g. ''she felt depressed'') may be included even when not explicitly diagnosed, potentially introducing inconsistencies in downstream generation.

These errors reflect inherent ambiguity in psychiatric narratives rather than systematic structural failures. Importantly, because perturbation and generation operate on validated graph structures, such inconsistencies remain localized and do not propagate into major diagnostic distortions. Additional examples and analysis are provided in Appendix~\ref{sec:app-err}.

\subsection{Summary}

Across both human and automated evaluations, Anonpsy achieves the intended combination of \textbf{clinical structure preservation} and \textbf{strong de-identification robustness}. Human psychiatrists and GPT-5 concur that Anonpsy narratives support DSM-5-TR diagnostic reasoning as well as the original text and a strong LLM-only baseline, yet are substantially less similar to and less recallable as the original patient. The LLM-only baseline preserves clinical information but insufficiently alters narrative identity, whereas Anonpsy introduces controlled, structure-aware transformations that significantly reduce re-identification risk without compromising downstream clinical utility.

\section{Conclusion}
We present Anonpsy, a graph-based framework for de-identifying psychiatric case histories by treating the task as a structure-preserving generation problem. By converting psychiatric narratives into semantic graphs, applying graph-constrained perturbations, and regenerating narratives through graph-conditioned text generation, our approach preserves the clinical structure necessary for downstream reasoning while substantially reducing re-identification risk. Experiments using human expert judgments and GPT-5 evaluations show that Anonpsy achieves consistently minimal re-identification risk, with risk scores concentrated at the lowest end of the evaluation scale, while yielding markedly lower semantic similarity and identifiability than strong LLM-only rewriting baselines. At the same time, clinical structure is preserved: both psychiatrist and GPT-5 evaluations detect no statistically significant degradation in clinical reasoning or diagnosis recoverability relative to the original narratives. Although instantiated for psychiatric narratives, the underlying paradigm of graph-based, structure-preserving de-identification may also be applicable to other narrative-style clinical documentation settings where identity is encoded through event content rather than explicit lexical identifiers.

\section*{Acknowledgments}

This research was supported by the Bio\&Medical Technology Development Program of the National Research Foundation (NRF) funded by the Korean government (MSIT; RS-2025-02263045), and by a grant from the Institute for AI and Social Innovation at Yonsei University (2025-22-0484).

\section*{Conflict of Interest}

B.-H. Kim serves as an outside director and holds equity in EverEx. This affiliation had no role in study design, data collection or analysis, decision to publish, or preparation of the manuscript.

\section*{Limitations}

Although Anonpsy demonstrates strong de-identification robustness while preserving clinical structure, several limitations remain. First, our evaluation relies on DSM-5-TR Clinical Cases, a curated educational corpus with coherent narrative flow and fully specified diagnostic labels. Real-world psychiatric documentation is more heterogeneous and varies widely in completeness and writing style, which may introduce additional challenges for semantic graph conversion and context-aware perturbation. In particular, extremely brief or fragmented clinical notes with limited narrative context or temporal structure may provide insufficient information for episode-level graph normalization, potentially requiring additional preprocessing or schema adaptation prior to applying the pipeline. Nevertheless, the underlying representation operates at the level of clinical entities and typed relations rather than corpus-specific formatting conventions, which facilitates adaptation across narrative-style psychiatric corpora. Evaluation on additional psychiatric corpora, such as publicly available psychiatric case collections or clinical narrative subsets, would further strengthen empirical validation and remains an important direction for future work.

Second, the human expert evaluation covers 20 cases due to the specialized expertise required. Although automated GPT-5-based evaluation on all 90 cases shows consistent trends, broader studies involving more clinicians and more diverse datasets would further strengthen generalizability.

Third, the perturbation strategies used in Anonpsy are currently based on clinically informed heuristic rules rather than a formally optimized objective. While these constraints are designed to preserve DSM-5-TR diagnostic structure, the broader design space of perturbation policies remains underexplored. Future work could formalize structure-preserving de-identification as a constrained optimization or invariance-preservation problem.

Fourth, our empirical comparison focuses on PHI masking and a carefully controlled LLM-only rewriting baseline. Although this design isolates the effect of explicit graph-based structural constraints, broader comparisons with additional de-identification systems or privacy-preserving text generation frameworks remain an important direction for future work.

Finally, while Anonpsy substantially reduces re-identification risk, residual risk may persist for cases involving rare diagnoses or highly distinctive symptom profiles, and de-identified narratives should not be treated as risk-free substitutes for real clinical records without additional safeguards.

\clearpage

\appendix
\section{Trade-off Premise Check Study}
\label{sec:tradeoff}
Figure~\ref{fig:schematic1} conceptualizes an implicit recallability--structure trade-off in psychiatric narrative de-identification. Because there is no single standardized, dataset-agnostic metric for real-world re-identification risk in free-text psychiatric narratives, we conduct a supporting analysis to \emph{operationalize} the two axes using automatic, reproducible proxies. This analysis is \emph{not} used as the main evaluation of privacy risk in our work; rather, it provides a concrete instantiation of the conceptual plane and allows us to situate common baselines and Anonpsy within a comparable space.

\subsection{Overview}
\label{sec:tradeoff_overview}
Given a set of original case narratives and corresponding de-identified variants produced by multiple methods, we quantify:
\begin{enumerate}
    \item \textbf{Narrative recallability} via semantic cosine similarity between the original narrative and each de-identified variant; and
    \item \textbf{Clinical structural preservation} via diagnosis preservation, quantified as a soft-F1 score of DSM-5-TR diagnoses predicted from each narrative variant relative to a fixed ground-truth diagnosis list.
\end{enumerate}

\subsection{Baselines and narrative variants}
\label{sec:tradeoff_variants}
We compute both proxies on the same set of cases across the following narrative sources:
\begin{itemize}
    \item \textbf{Anonpsy}: our graph-guided de-identification output.
    \item \textbf{PHI masking}: a token-level masking baseline that replaces surface identifiers using a combination of regex rules and named entity recognition (spaCy).
    \item \textbf{SDC}: an unconstrained one-pass synthetic rewriting baseline produced by a local LLM (\texttt{gpt-oss:120b}).
\end{itemize}

\subsection{Narrative recallability proxy: semantic cosine similarity}
\label{sec:tradeoff_cosine}
We operationalize narrative recallability by calculating the cosine similarity for all methods used to populate the recallability--structure plane (Anonpsy, PHI masking, and SDC). If a de-identification method leaves idiosyncratic narrative content largely intact, the resulting variant will exhibit higher cosine similarity to the original due to greater semantic overlap. Conversely, methods that substantially perturb personal narrative content are expected to yield lower cosine similarity, reflecting reduced narrative recallability. We use the same semantic cosine similarity procedure described in Appendix~\ref{sec:cosine}.

\subsection{Structure proxy: diagnosis preservation via soft-F1}
\label{sec:tradeoff_softf1}
We operationalize structural fidelity by measuring whether each narrative variant preserves diagnostically essential information sufficient for accurate DSM-5-TR read-out. If de-identification distorts symptom trajectories or key contextual dependencies, diagnostic predictions should degrade; conversely, variants that preserve diagnostic logic should sustain performance. We use the same GPT-5 diagnosis prediction and soft-F1 evaluation pipeline described in Appendix~\ref{sec:gpt:diag:pred}.

\subsection{Trade-off Verification Results and Interpretation}
\label{sec:tradeoff_results}

\paragraph{Narrative recallability (semantic cosine similarity).}
Semantic cosine similarity results revealed strong separation between methods. PHI masking exhibited the highest similarity to the original narratives (mean = 0.902), reflecting extensive semantic overlap. SDC showed moderately reduced similarity (mean = 0.796), with Anonpsy occupying the lower end of the similarity range (mean = 0.718).

\paragraph{Structure preservation (diagnosis soft-F1).}
Using soft-F1 as a proxy for clinical structural fidelity, the original narratives achieved a mean score of 0.665 (SD = 0.276). Among de-identified variants, PHI masking retained the highest diagnostic fidelity (mean = 0.618, SD = 0.316), followed by Anonpsy (mean = 0.567, SD = 0.316) and SDC (mean = 0.554, SD = 0.346).

\paragraph{Interpretation.}
Taken together, these findings empirically instantiate the conceptual recallability--structure trade-off illustrated in Figure~\ref{fig:schematic1}. PHI masking preserves diagnostic structure but retains high narrative recallability, leaving residual re-identification risk. SDC reduces recallability to some extent but does so at the cost of diagnostically essential structure. Anonpsy occupies a distinct region of the trade-off space, combining substantially reduced narrative recallability with preserved diagnostic structure. Although this analysis serves as a supporting verification rather than a primary privacy evaluation, it provides a quantitative illustration of how structured, graph-guided de-identification can balance privacy protection and clinical utility relative to existing unstructured approaches.

\section{Methodological Details}
\label{sec:method_detail}
\subsection{Text-to-Graph Conversion}
\label{sec:conversion}
\subsubsection{Conversion pipeline full schema}
Conversion from free-text narrative to a semantic graph is performed using a staged, schema-constrained LLM-based pipeline, corresponding to the conversion operator $\mathcal{E}$. Rather than relying on a single end-to-end extraction step, the conversion process is decomposed into four conceptual stages: (1) schema-constrained clinical entity extraction, (2) temporal episode construction and normalization, (3) temporal reconciliation and node canonicalization, and (4) typed relation construction and graph finalization. LLMs are employed for schema-guided extraction and constrained semantic inference, while temporal normalization, reconciliation, and consistency checks are implemented deterministically. This staged organization enables intermediate validation, enforces schema constraints, and limits error propagation across components.

\paragraph{Schema-constrained clinical entity extraction.}
Clinical entities are first identified under explicit schema constraints. Global attributes that do not participate in relational structure are extracted at the case level, including demographics (age, sex, ethnicity, occupation, and family structure), family history, and objective test results. Test results are organized into four case-level fields: \texttt{labs}, \texttt{imaging}, \texttt{mental\_status}, and \texttt{other}. Laboratory and imaging results are stored as semi-structured summaries, while the mental status examination is preserved as a single near-verbatim string to avoid over-fragmentation of clinically cohesive observations; remaining objective assessments such as psychiatric scales or physical examination findings are grouped under \texttt{other}.

Core graph nodes are then instantiated. Diagnosis nodes are created directly from diagnostic metadata provided in the input data. Symptom nodes are extracted as DSM-aligned headwords with near-verbatim \texttt{evidence\_text} and are provisionally aligned to diagnoses under constrained decoding. Contextual descriptors associated with each symptom are decomposed using the STEB schema, with each field restricted to patient-attributable narrative content rather than clinician interpretation or test data. An initial \texttt{current\_symptom} flag is assigned at this stage and later recomputed deterministically. Treatment nodes are extracted with schema-limited attributes (type, name, dose, route, frequency, and outcome), with route values normalized to a controlled vocabulary and populated only for administrable interventions. A single \texttt{VisitEvent} node is instantiated to represent the index encounter, encoding setting, arrival mode, legal status, reason for visit, safety flags, and source of information, along with a short near-verbatim \texttt{visit\_episode} anchoring the presentation narrative. Past-history nodes are created from explicit past-history mentions and diagnoses marked as pre-existing or remitted.

\paragraph{Temporal episode construction and normalization.}
For each symptom, treatment, and past-history item, one or more temporal episodes relative to the index encounter (day~0) are extracted. Episodes are initially represented as (\texttt{offset}, \texttt{span}, \texttt{unit}) triples with an optional \texttt{ongoing} flag. When explicit temporal expressions are present, they are parsed directly; when timing is coarse or implicit, such as  "for several years," "in childhood," or "recently," expressions are mapped to the coarsest compatible offset and span supported by the text.

If no explicit temporal expression is provided, a constrained, order-preserving inference is applied: relative position within the narrative and relations to other temporally grounded events are used to assign a plausible relative offset, without introducing absolute dates or fine-grained spans. This inference is conservative and respects narrative order; when insufficient evidence exists to localize an event beyond ordering, temporal boundaries are left underspecified.

When an episode is ongoing or lacks an explicit end, a finite span is deterministically assigned using a shared timeline horizon derived from other episodes. All episodes are then canonicalized into a unified day-based coordinate system, yielding half-open intervals $[start, end)$ with $start=\texttt{offset}$ and $end=\texttt{offset}+\texttt{span}$. Episode signatures are globally deduplicated by exact match on their day-based representation.

\paragraph{Temporal reconciliation and node canonicalization.}
After global deduplication, overlapping or adjacent intervals referenced by the same node are reconciled and merged into disjoint blocks. This reconciliation is implemented by creating node-specific virtual duration objects and rewriting each node’s \texttt{duration\_ids} to reference the merged intervals; the process is iterated until a fixed point is reached, and unreferenced durations are removed. The \texttt{current\_symptom} flag is then recomputed deterministically: a symptom is labeled current if any reconciled interval covers day~0 under the half-open interval rule. To enforce a one-to-one correspondence between symptom nodes and temporal intervals, symptoms associated with multiple distinct episodes are split into one-episode-per-node representations, with contextual STEB frames paired to episodes by temporal order. This stage fixes the temporal backbone of the graph and canonically defines node identity.

\subsubsection{Typed relation construction and graph finalization.}
Once node inventories and temporal structure are fixed, typed inter-entity relations are constructed using a combination of deterministic rules and constrained model-based selection. These include diagnostic links (\texttt{MANIFESTS\_AS}), therapeutic links (\texttt{TREATMENT\_OF}) to diagnoses or past-history targets, and visit-level presentation links (\texttt{PRESENTS\_WITH}) from the \texttt{VisitEvent} node to all symptoms whose reconciled temporal intervals cover day~0.

Etiologic relations (\texttt{INDUCES}) are added conservatively in two steps. First, diagnosis labels containing explicit etiologic morphology (e.g., "due to," "-induced") are parsed and linked to the most appropriate anchor. Second, an additional constrained LLM-based causal pass introduces further \texttt{INDUCES} edges only when explicit causal statements are supported by near-verbatim textual evidence.

After all stages, the graph undergoes final consistency checks and is serialized into a stable YAML representation consisting of typed nodes, typed relations, and a shared pool of normalized duration objects. This semantic graph serves as the sole input to downstream perturbation and graph-conditioned text generation.

\subsubsection{Formal semantic graph schema}
This section describes the structure of the semantic graph produced during conversion from free-text psychiatric narratives. The representation is designed to make clinically meaningful structure explicit while supporting targeted perturbation of identifying content. An overview of the formal semantic graph schema is shown in Table~\ref{tab:schema_overview}.

\begin{table*}[t]
\centering
\small
\begin{tabular}{p{3.2cm} p{3.0cm} p{7.5cm}}
\toprule
\textbf{Component} & \textbf{YAML key} & \textbf{Description} \\
\midrule
\textbf{Case-level attributes} & \texttt{demographics} &
Age, sex, ethnicity, occupation. \\

& \texttt{test\_results} &
Objective findings grouped under \texttt{labs}, \texttt{imaging}, \texttt{mental\_status}, \texttt{other}. \\

& \texttt{family\_history} &
Family-history entries with member, condition, and supporting evidence text. \\

\midrule
\textbf{Graph nodes} & \texttt{diagnoses} &
DSM-5-TR diagnoses associated with the case. \\

& \texttt{symptoms} &
DSM-aligned symptom headwords, each associated with one or more duration-based episodes in STEB schema. \\

& \texttt{treatments} &
Medications, psychotherapies, or procedures, represented with attributes such as name, dose, route, frequency, and outcome. \\

& \texttt{past\_history} &
Prior psychiatric or medical conditions, including chronic and remitted states. \\

& \texttt{visit\_event} &
A single node representing the index clinical encounter, encoding setting, arrival mode, legal status, reason for visit, safety flags, and source of information. \\

\midrule
\textbf{Temporal objects} & \texttt{durations} &
Shared pool of normalized intervals (\texttt{offset}, \texttt{span}, \texttt{unit}) in days relative to day~0. \\

\midrule
\textbf{Typed relations} & \texttt{MANIFESTS\_AS} &
Links a symptom to an associated diagnosis. \\

& \texttt{TREATMENT\_OF} & 
Links a treatment to its diagnostic target. \\

&\texttt{PRESENTS\_WITH} &
Links the  \texttt{VisitEvent} node to symptoms active at day~0.\\

& \texttt{INDUCES} &
Encodes etiologic relationships between entities when explicit causal structure is present. \\

\bottomrule
\end{tabular}
\caption{Schema overview of the semantic graph. Case-level attributes are stored outside the relational graph; graph structure is represented via typed nodes, typed relations, and shared temporal duration objects.}
\label{tab:schema_overview}
\end{table*}

\paragraph{Symptom context representation (STEB)}
Each symptom node contains one or more contextual frames decomposed into the STEB schema:
\begin{itemize}
    \item \textbf{Situation}: an external or observable context.
    \item \textbf{Thought}: the patient’s internal appraisal or interpretation of the situation.
    \item \textbf{Emotion}: the affective state associated with the situation.
    \item \textbf{Behavior}: the observable behavioral response to the situation.
\end{itemize}
The STEB schema serves as a modular scaffold that isolates manipulable narrative components while preserving clinically meaningful structure. To ensure that this decomposition is psychiatrically grounded and supports controlled perturbation, the schema is informed by the cognitive formulation framework commonly used in cognitive-behavioral therapy \citep{beck1979cognitive}.

\paragraph{Temporal representation}
Temporal information is encoded via a shared pool of duration objects referenced by node identifiers. Each duration represents a half-open interval $[start, end)$ in integer days relative to the index encounter (day~0), where negative offsets denote pre-encounter episodes and positive offsets denote post-encounter episodes. Each symptom node references exactly one reconciled duration object. A symptom is labeled \texttt{current\_symptom} if its associated interval covers day~0 under the half-open rule.

\paragraph{Illustrative serialization example}
\label{serial}
The semantic graph is serialized to a transparent YAML format. The following synthetic example illustrates how a single symptom node is represented, including its STEB context and associated temporal reference.

\begin{lstlisting}
- id: s_003
  symptom: ideas of reference
  pattern: continuous
  current_symptom: true
  evidence_text: the news announcers began to comment indirectly and critically
    about him.
  contexts:
    - situation: while watching a late-night news program
    - thought: the news announcers were commenting indirectly and critically about me.
    - emotion: anxious
    - behavior: repeatedly called the television station to complain about the broadcast.
  duration_ids: [dvm_048]
\end{lstlisting}

\subsection{Graph-constrained Perturbation}
\label{sec:perturb}
\subsubsection{Demographics perturbation.}
Quasi-identifying demographic attributes (age, sex, ethnicity, occupation) are perturbed using rule-based constraints combined with LLM-guided substitutions. Age is modified through a bounded offset procedure that adjusts all age-referenced durations consistently and enforces DSM-5-TR feasibility rules for both present age and age of onset. For example, antisocial personality disorder requires age~$\geq 18$ and neurodevelopmental disorders require childhood onset. Sex is perturbed using a weighted-random scheme with hard constraints for sex-specific diagnoses or labels. Ethnicity and occupation are rewritten through LLM prompts that generate plausible but non-identifying alternatives while maintaining functional realism; for example, we restrict job types for minors and preserve support-level requirements in neurodevelopmental disorders. These perturbed demographic attributes propagate to the graph-conditioned text generation module but do not affect the graph’s relational structure, as demographic fields do not participate in typed edges.

\subsubsection{Symptom-context perturbation.}
Narrative perturbation operates in two stages: rewriting the visit-event episode and then perturbing the contextual descriptors associated with each symptom node. Both stages preserve the clinical structure encoded in the graph.

\paragraph{Visit-event rewriting.}
From the structured \texttt{visit\_event} fields, we derive a fixed scaffold capturing the core clinical attributes of the presentation—voluntary versus involuntary status, arrival mode, clinical setting, and urgency inferred from safety flags. These attributes are treated as immutable constraints. The LLM must generate a rewritten visit-episode narrative that remains fully compatible with this scaffold, ensuring that clinically defining properties do not change. In other words, a voluntary outpatient visit accompanied by family cannot become an involuntary police-escorted emergency arrival. Within these constraints, the LLM freely rewrites the narrative details surrounding the presentation and removes identifying elements such as names, locations, addresses, and specific dates. A secondary pass optionally rewrites the concise pathway field when the original phrasing is overly specific.

\paragraph{Symptom-context rewriting.}
Each symptom node contains one or more contextual descriptions in the STEB schema. Only the STEB fields present in the original node are rewritten; no new fields are introduced. For each context, an LLM generates a replacement that preserves the symptom’s clinical meaning while substituting identifying psychosocial details with non-specific or semantically analogous alternatives. Temporal offsets, duration references, node identifiers, and diagnostic associations remain unchanged. Rewrites incorporate the patient’s age at the event, computed from duration offsets and the perturbed demographics, to maintain developmental and clinical plausibility.

All symptom contexts are perturbed in a chronologically retrograde order. This ensures that the rewritten visit episode anchors the most recent clinical scenes, and each successively earlier symptom context is rewritten with access to a short window of the already edited contexts. This retrograde schedule improves narrative coherence by allowing later, clinically salient events—such as the circumstances of presentation and current symptomatology—to shape the rewrites of earlier episodes without introducing forward-looking inconsistencies. After each rewrite, a similarity filter rejects outputs that are too close to the original text, triggering controlled LLM retries.

Together, the scaffold-constrained visit-event rewrite and STEB-based
symptom-context perturbations produce a coherent yet privacy-preserving
narrative that preserves clinical content while removing or altering potentially
identifying contextual details.

\subsubsection{Test-results perturbation.}
In addition to demographic and symptom-context editing, we perturb numeric test results appearing in the \texttt{test\_results} fields of the YAML file. We extract numeric values from free-text test summaries and map each extracted entry to a canonical test name drawn from a curated inventory (for example, Full Scale IQ, MMSE, or fasting glucose). Each canonical test is associated with clinically meaningful value pools representing ranges that preserve interpretability. The system identifies the pool containing the original value and resamples a new value from within the same pool, ensuring that the rewritten test remains clinically equivalent while eliminating exact numeric matches that could serve as linkage identifiers. The substitution is performed minimally within the text, maintaining surrounding phrasing and narrative context. Because only numeric values are altered and the clinical category of each test remains unchanged, this procedure reduces re-identification risk while preserving all downstream diagnostic implications.

\paragraph{Mental Status Examination test results.}
The Mental Status Examination (MSE) may contain descriptors that implicitly depend on demographics or on themes expressed in symptom thoughts. Following demographic and symptom-context perturbation, we therefore apply a targeted alignment. The module compares the pre- and post-perturbation essence objects to detect changes in age, sex, ethnicity, occupation, or visit-episode framing, and collects a small set of rewritten symptom thoughts to capture the updated thematic landscape. A constrained LLM prompt then produces a minimally edited MSE paragraph that harmonizes these fields, adjusting only those phrases that depend on demographic or thematic consistency while preserving all objective findings related to appearance, speech, mood and affect, thought process and content, perception, orientation, insight, and judgment. The editor is not permitted to introduce new symptoms or delete clinically relevant features. This alignment step ensures that the perturbed case history remains narratively coherent and clinically faithful in its final form.

\subsubsection{Structural and temporal consistency enforcement.}
Following attribute perturbation, the system re-validates graph consistency. Temporal constraints, including duration spans, onset offsets, and episode overlaps, are inherited unchanged from $G$; perturbations are not permitted to alter the temporal backbone. Similarly, relational constraints are preserved: if a treatment targeted a specific symptom or diagnosis in $G$, the same \texttt{TREATMENT\_OF} edge must remain valid in $\tilde{G}$, and all node-type pairings remain schema-authorized. A consistency checker verifies that \[\text{Temporal}(v_i, v_j; G) = \text{Temporal}(v_i, v_j; \tilde{G})\] for all clinically constrained event pairs, and that perturbations do not introduce contradictory or unsupported causal or diagnostic relations.

This graph-based perturbation procedure ensures that re-identification risks are reduced while maintaining the clinical logic, symptom trajectory, and relational structure of the original case.

\subsection{Graph-conditioned Text Generation}
\label{sec:textgen}
\subsubsection{Overview.}
Given a perturbed semantic graph $\tilde{G}$, the graph-to-text operator $\mathcal{D}$ constructs a de-identified case history in three stages: (1) a lead paragraph describing the index presentation; (2) a deterministic chronological history of present illness built from graph structure; and (3) a short appendix that summarizes remaining past history, family history, and day~0 test results.

\subsubsection{Lead paragraph generation.}
The first step produces a short lead paragraph that introduces the patient and
the index encounter. We condition a ChatOllama instance (\texttt{gpt-oss:120b}, temperature 0.1) on a compact JSON summary containing age, sex, care setting, arrival mode, reason for visit, pathway, source of history, and a brief visit-episode description. The system prompt instructs the model to write two to five sentences in third-person past tense, covering who, where, and why, and explicitly forbids identifiers such as names or addresses. The model returns only the paragraph, which is post-processed to strip meta-text.

\subsubsection{Chronological history from the graph.}
The second stage builds a deterministic, graph-driven history. We first derive durations (with day offsets and spans), the mapping from each symptom and treatment to its duration episodes, diagnosis labels, and the various relation-type maps. Durations are sorted by \texttt{offset\_days} to define a global timeline from earliest to latest event. A helper converts each offset into a human-readable time phrase such as “two weeks before admission” or “the day after admission,” with the lexicon switching automatically between ``before/after admission'' and ``earlier/later'' depending on whether the case contains clear admission-type events.

For each duration block, we group symptoms by their linked diagnosis and generate micro-paragraphs per (duration, diagnosis) pair. The generator walks through:
\begin{enumerate}
    \item \textbf{Symptom narration.}
    For each symptom, we retrieve its STEB context and construct a single
    sentence that blends the time phrase with situation, thought, emotion, and
    behavior. We first call a one-sentence prompt that asks the model to depict a concrete episode of the symptom using the STEB fields (temperature 0.2). A bookkeeping object ensures that each (item, duration) pair is narrated at most once.
    \item \textbf{Treatment narration and co-regimens.}
    For each diagnosis and duration, we attach treatments that target that diagnosis and share the same duration. When treatments are narrated, additional co-treatments that share both duration and target diagnosis are listed as part of the same regimen sentence.
    \item \textbf{INDUCES traversal.}
    If a symptom, treatment, or past-history node has an \texttt{INDUCES} relation to a diagnosis, we traverse that edge and narrate a short cluster of induced symptoms and treatments anchored to the same temporal context, ensuring that etiologic chains (for example, “steroid-induced psychosis”) are explicitly represented.
\end{enumerate}

Past-history nodes are handled in a preliminary pass: each past condition is
narrated once at its earliest known offset if it either induces a diagnosis or
is linked to a treatment episode, optionally followed by sentences describing
treatments for that condition. Past-history nodes which are not linked to any other nodes are appended later in the “Appending past history, family history, and tests” step.

\subsubsection{Appending past history, family history, and tests.}
After constructing the lead and chronological history, we compute which past-history diagnoses have not yet been mentioned, collect family-history summaries from the case-level attributes, and aggregate day~0 test results. For tests, we combine essence-level results from the \texttt{labs}, \texttt{imaging}, \texttt{mental\_status}, and \texttt{other} fields.

We pass the draft narrative plus these structured lists to a ChatOllama prompt that appends one to four sentences at the end of the case. The prompt explicitly instructs the model to: (1) add a compact sentence on remaining past history; (2) add a compact sentence on family history; and (3) add a clause-dense sentence summarizing day~0 objective test data. Privacy rules are reiterated, and the model may not restate the lead or rewrite prior sentences.

\section{Hyperparameter Settings}
\label{sec:hyperparams}

Table~\ref{tab:hyperparams} summarizes the decoding hyperparameters used for the LLM-assisted components of the Anonpsy pipeline as well as the LLM-only baseline. Hyperparameters are grouped by operator ($\mathcal{E}$, $\mathcal{P}$, $\mathcal{D}$) and reflect fixed settings used across all experiments unless otherwise noted. The LLM-only baseline employs a non-zero temperature for the constrained clinical rewrite step and a deterministic temperature for the self-critique and refinement step.

\begin{table}[t]
\centering
\small
\begin{tabular}{lc}
\toprule
\textbf{Component} & \textbf{Temperature} \\
\midrule
Conversion operator ($\mathcal{E}$)        & 0.1 \\
Perturbation operator ($\mathcal{P}$)      & 0.7 \\
Generation operator ($\mathcal{D}$)        & 0.1 \\
\midrule
LLM-only (Constrained Clinical Rewrite)             & 0.2 \\
LLM-only (Self-Critique and Refinement)               & 0.0 \\
\bottomrule
\end{tabular}
\caption{Temperature settings used for all LLM-assisted components.}
\label{tab:hyperparams}
\end{table}

\section{Experimental Details}
\label{sec:exp-details}
\subsection{LLM-only Baseline Construction}
\label{sec:llmonly}
To construct a strong text-only baseline, we used the same LLM as Anonpsy (i.e., gpt-oss:120b) but without access to the graph structure. Rewriting proceeded in two stages. Stage~1 applied a constrained clinical rewrite prompt that (i) preserved all diagnoses and their supporting evidence, (ii) preserved temporal and causal relations among symptoms, and (iii) maintained severity patterns and developmental trajectories, while (iv) requiring systematic modification of age, sex, occupation, locations, interpersonal details, quotations, psychometric values, dates, proper nouns, and institutions. The prompt also required the output to follow the same macro-structure as Anonpsy (lead presentation, chronological history of symptoms, concluding past history, family history, and test results) to control for format as a confound. Stage~2 applied a self-critique refinement prompt to the Stage-1 output, with explicit instructions to remove residual copied idiosyncratic events and to repair any privacy- or fidelity-violating content. This two-stage setup yields a strong, structurally controlled LLM-only baseline. Full prompt is presented below. Exact decoding temperature settings are reported in Appendix~\ref{sec:hyperparams}, Table~\ref{tab:hyperparams}.

\subsubsection{LLM-only prompts.}
\paragraph{Stage 1: Constrained Clinical Rewrite Prompt}
\begin{lstlisting}
SYSTEM:
"You are de-identifying psychiatric case narratives while preserving clinical
fidelity. Be concise, coherent, and clinically plausible. Avoid stylistic
verbosity."

USER:
You are rewriting the following psychiatric case history for educational use.
Your goal is to produce a *new narrative* that preserves all core clinical
findings and diagnostic reasoning, but changes the surface details so that no
reader could reasonably identify the original patient.

INSTRUCTIONS:
- Preserve:
  * Diagnoses and their justifying clinical features
  * Temporal and causal relations among symptoms
  * Severity levels and developmental trajectories

- Alter:
  * Demographics (e.g., name, age, sex, occupation)
  * Locations, family structures, interpersonal details
  * Idiosyncratic life events and hobbies (replace with analogous but distinct ones)
  * Verbatim quotes or distinctive phrasing (rephrase)
  * Psychometric scores, ages, milestones (generalize or jitter plausibly)
  * Specific institutions or proper nouns (replace with generic equivalents)

- Maintain coherent chronology, tone, and style.
- Output should read as a plausible but *distinctly different life story*.

STRUCTURE REQUIREMENTS:
- Begin with the presenting reason for visit.
- Provide a chronological history of symptoms.
- Conclude with past psychiatric/medical history, family history,
  and test or examination findings.
- Do not introduce new diagnoses or analytic commentary.
- Write in paragraph form only.

Return only the rewritten case narrative.

--- INPUT ---
{case_text}
\end{lstlisting}

\paragraph{Stage 2: Self-Critique and Refinement Prompt}
\begin{lstlisting}
You are auditing the rewritten case for two criteria:

1. Privacy robustness:
   - Does any rare or unique detail appear copied from the original?
   - If so, modify while preserving plausibility.

2. Clinical fidelity:
   - Are diagnoses, symptom patterns, and severity preserved?
   - Is the chronology coherent and medically reasonable?

Make minimal necessary edits. Return the final anonymized narrative only.

--- DRAFT ---
{draft_text}
\end{lstlisting}

\subsection{Human Clinical Structure Study Protocol and Analysis}
\label{sec:human_clinical}
In the clinical structure experiment, each psychiatrist rated 10 cases, each presented in all three narrative variants in randomized order. Raters were told that the task was to evaluate the clinical validity of synthetic psychiatric case histories; they were not informed that multiple versions of the same underlying case existed. For each narrative, raters answered Q1 (open-ended DSM-5-TR diagnosis) and Q2 (binary diagnosis-narrative congruence).

Q1 responses were normalized by lowercasing and whitespace normalization but otherwise preserved DSM surface forms, including specifiers, to test whether fine-grained diagnostic cues were retained. Because each of the five psychiatrists evaluated the same 10 cases, the resulting 150 observations exhibit two structured dependencies:  
(1) \emph{case difficulty}, where all raters tend to score some cases higher or lower depending on their inherent diagnostic complexity; and  
(2) \emph{rater diagnostic style}, reflecting stable individual tendencies such as stricter or more lenient thresholds for comorbid diagnoses.  
To account for these hierarchical dependencies, we fit a linear mixed-effects model,
\[
\text{soft\_F1} \sim \text{C(input\_type)} + (1 \mid \text{case}) + (1 \mid \text{rater}),
\]
treating \emph{Anonpsy} as the reference level. This approach allows the analysis to use all observations while correctly attributing variance to case-level and rater-level effects rather than confounding them with the fixed effect of interest (input type). As a non-parametric complement, we also computed per-(case,input) means and conducted Wilcoxon signed-rank tests (Table~\ref{tab:q1_mixed_model} in the main text).

For Q2 (narrative--diagnosis congruence), responses were binary and strongly imbalanced (90-96\% "yes" across conditions). Under such imbalance, classical logistic mixed-effects models can produce unstable maximum-likelihood estimates. We therefore fit a Bayesian logistic mixed-effects model,

{\small
\[
\text{logit}\, P(\text{Q2\_yes}=1)
   \sim \text{C(input\_type)} + (1 \mid \text{case}) + (1 \mid \text{rater}),
\]
}

with variance components for both case and rater. The Bayesian formulation yields stable posterior distributions and transparent uncertainty estimates for the effect of narrative type after adjusting for case difficulty and rater-specific response patterns. 

\subsection{Human Re-identification Study Protocol and Rubric}
\label{sec:rubric}
For the human re-identification experiment, each psychiatrist rated a disjoint set of 10 cases. For each case, they first read the original narrative, then an A/B pair of de-identified versions in randomized order, label concealed. Raters (i) selected which version felt more similar to the original and (ii) assigned 1--5 re-identification risk scores to both A and B using a structured rubric. The rubric, summarized in Table~\ref{tab:rubric}, distinguishes narratives that share only generic clinical features and are not realistically recallable (scores 1--2) from narratives that a knowledgeable insider could plausibly link back to a specific individual (scores $\geq 3$), based on distinctive narrative overlap and concrete anchors such as rare events, precise location or dates, or tightly characteristic life stories. This design yields paired risk scores for each method under a shared case and rater, controlling for case difficulty and individual rater stringency. We applied a two-sided Wilcoxon signed-rank test to the paired risk scores and a binomial test to the proportion of times LLM-only was chosen as more similar to the original. As descriptive supplements, we compared risk scores separately for chosen versus non-chosen versions using Mann--Whitney U tests and visualized full score distributions with violin plots (Figure~\ref{fig:risk_violin}).

\begin{table*}[t]

    \centering
    \small
    \begin{tabular}{p{2.5cm} p{12.5cm}}
        \hline
        \textbf{Score} & \textbf{Interpretation} \\
        \hline
        \textbf{1 (No risk)} &
        Episodes share only generic clinical features; narrative is clearly distinct from the original. \\[4pt]

        \textbf{2 (Minimal risk)} &
        Some thematic overlap but insufficient detail to identify the individual patient. \\[4pt]

        \textbf{3 (Moderate risk)} &
        Noticeable overlap in narrative content; a knowledgeable insider \emph{might} suspect it is the same patient despite the absence of explicit identifiers. \\[4pt]

        \textbf{4 (High risk)} &
        Score-3 level overlap plus at least one concrete anchor (exact dates, doses, places, rare events). \\[4pt]

        \textbf{5 (Very high risk)} &
        Paraphrased version of the same story containing multiple anchors. \\
        \hline
    \end{tabular}
    \caption{
    Re-identification risk rubric used in both GPT-5 and human evaluation.  
    }
\label{tab:rubric}
\end{table*}

\subsection{Qualitative Analysis: Narrative-Level Comparison}
\label{app:qualitative-example}

To complement the quantitative re-identification analysis, we present two illustrative cases from the human re-identification study. Together, these cases highlight both the strengths and the limitations of Anonpsy relative to a strong LLM-only baseline. The first case exemplifies a setting in which Anonpsy substantially reduces narrative identifiability while preserving psychiatric structure. The second illustrates a challenging failure mode in which both approaches exhibit residual recallability, albeit for different reasons.

In the first case, all five psychiatrists who evaluated the narratives judged the \emph{LLM-only} version as more recallable and at higher re-identification risk than the corresponding \emph{Anonpsy} output. Although explicit identifiers were removed, evaluators noted that the LLM-only output closely mirrored the original case trajectory. Comments emphasized substantial overlap in treatment course and symptom progression, including the sequence of antidepressant non-response followed by clomipramine augmentation, concurrent hypertension treated with propranolol, specific medication dosages, and the emergence of sexual dysfunction as a treatment side effect. One rater commented that the narrative "felt largely overlapping" (risk score 4), while another noted that "the course and symptom pattern are described in a very similar way" (risk score 4), characterizing the text as "more of a paraphrase" (risk score 3). 

This overlap was evident in passages such as the LLM-only description of treatment response: \emph{"After about five weeks of this regimen the patient reported substantial improvement \dots however, despite this renewed desire, he experienced repeated unsuccessful attempts at intercourse."} Evaluators reported that such phrasing closely tracked the original narrative’s structure and sequencing, contributing to recallability.

In contrast, the Anonpsy output retained the clinically essential structure of the case—recurrent major depressive disorder, severe psychomotor retardation, pharmacologic non-response, subsequent improvement, and treatment-emergent sexual dysfunction—while substantially altering its narrative realization. Demographic attributes, occupational context, and presentation framing were changed, and symptom severity was conveyed through embodied behavioral observations rather than replicating the original symptom listing. For example, sexual dysfunction was preserved as a clinically relevant outcome but embedded in a distinct emotional and interpersonal context, as illustrated by phrasing such as \emph{"he expressed fear that the problem might be permanent, avoiding eye contact and repeatedly voicing shame,"} rather than a medication-centered side-effect report. Evaluators consistently reported that any perceived similarity in the Anonpsy narrative arose from generic psychiatric patterns rather than distinctive narrative anchors, noting that the case was "coherent but not recognizable as the same person."

The second case represents a challenging scenario involving chronic obsessive-compulsive disorder dominated by contamination obsessions and extensive cleaning rituals. In this case, psychiatrists reported that both Anonpsy and the LLM-only baseline evoked the original patient, indicating a shared failure mode. However, evaluators emphasized that the mechanisms of residual recallability differed between the two approaches. For the LLM-only version, recallability was attributed to the persistence of concrete narrative anchors, including pandemic-related anecdotes and specific cleaning behaviors such as bleach use. One rater explicitly noted that "COVID-19-related episodes and bleach usage appear to function as identifiable cues"(risk score 3).

For the Anonpsy version, by contrast, evaluators observed that recognizability arose not from copied events or contextual anchors, but from the preservation of a highly faithful and coherent depiction of psychopathology. Despite altered demographics and narrative framing, the temporally grounded portrayal of compulsive rituals, intrusive contamination fears, and retained insight into their irrationality remained strongly evocative. One psychiatrist commented that ``the description is coherent and vivid, and because obsessive--compulsive symptoms dominate the narrative, the patient comes to mind even though the details differ'' (risk score 2). Another noted that ``demographic features are similar, but the clinical presentation is described differently'' (risk score 2). Notably, although these evaluators selected Anonpsy as relatively more recallable in this case, both assigned a risk score of 2, corresponding to \emph{minimal risk} under the rubric, indicating thematic overlap without sufficient detail to identify the individual patient.

Taken together, these two cases clarify both the strengths and the boundary conditions of Anonpsy. In many settings, graph-constrained perturbation effectively suppresses narrative identity while preserving clinically essential temporal, causal, and diagnostic structure, outperforming LLM-only rewriting that inadvertently retains illness trajectories and treatment sequences. However, in disorders where identity is tightly coupled to symptomatology itself, such as severe obsessive--compulsive disorder, reducing re-identification risk remains challenging for both approaches. Importantly, even in such cases, the mechanisms of residual recallability differ: LLM-only preserves concrete narrative anchors, whereas Anonpsy preserves a rich and coherent symptom structure. 

\subsection{GPT-5 Diagnosis Prediction Accuracy}
\label{sec:gpt:diag:pred}
For diagnosis prediction, we assessed whether each narrative variant enabled GPT-5 to infer the correct DSM-5-TR diagnoses. For each case, GPT-5 received only the narrative and generated a diagnosis list, which we evaluated using a soft-F1 metric based on greedy semantic matching. Because biomedical text typically exhibits substantial synonymy and surface-form variation, raw string matches can incorrectly penalize predictions that are clinically equivalent \citep{kartchner2023comprehensive, leaman2015challenges}. In the psychiatric domain, this problem is amplified: DSM-5-TR diagnostic labels contain numerous surface-level variations that do not meaningfully alter the underlying disorder. Many labels include specifiers (e.g., \emph{in remission}, \emph{most recent episode depressed}, \emph{with psychotic features}) that modify course or severity but do not change the base diagnosis. Furthermore, DSM-5-TR coding rules require substituting the specific substance, medication, or medical condition into umbrella categories such as diagnosing \emph{sedative, hypnotic, or anxiolytic use disorder} as \emph{alprazolam use disorder}. When LLM returns the umbrella category as a diagnosis, this does not necessarily indicate loss of clinical structure in the input narrative. To ensure a clinically meaningful and fair comparison, we therefore performed canonicalization: including lowercasing, removal of specifiers, and mapping known DSM synonyms (e.g., \emph{conversion disorder} and \emph{functional neurological symptom disorder}) to a shared canonical form. This procedure ensures that evaluation reflects true diagnostic preservation rather than artifacts of DSM naming conventions or LLM phrasing. 

\subsubsection{Results.}  
Table~\ref{tab:diag_f1} presents descriptive statistics and significance tests across the 90 paired cases. Although mean soft-F1 differs across conditions, pairwise Wilcoxon tests with Holm correction show no statistically significant difference.

\subsubsection{Interpretation.}  
These results indicate that \textbf{Anonpsy preserves diagnostic usability}.  
Its outputs maintain the clinical structure needed for GPT-5 to recover DSM-5-TR diagnoses with accuracy comparable to both the original text and LLM-only baseline

\begin{table}[t]
\centering
\small
\begin{tabular}{lccc}
\toprule
\textbf{Group} & \textbf{Mean F1} & \textbf{Median} & \textbf{SD} \\
\midrule
Original & 0.644 & 0.667 & 0.333 \\
Anonpsy & 0.567 & 0.600 & 0.316 \\
LLM-only & 0.662 & 0.667 & 0.329 \\
\bottomrule
\end{tabular}

\vspace{3pt}
\begin{tabular}{lcc}
\multicolumn{3}{c}{\textbf{Statistical Tests}} \\
\toprule
Friedman $\chi^2$ & 15.174 & $p < 0.001$ \\
\midrule
Pairwise Wilcoxon (Holm) &  &  \\
Anonpsy vs.\ LLM-only & & 0.997 \\
Anonpsy vs.\ Original & & 1.000 \\
LLM-only vs.\ Original & & 0.602 \\
\bottomrule
\end{tabular}

\caption{Diagnosis prediction accuracy (soft-F1) across the three narrative variants.}
\label{tab:diag_f1}
\end{table}

\subsection{GPT-5 Diagnosis Acceptability}
\label{sec:gpt:diag:acc}
For diagnosis acceptability, GPT-5 received both the ground-truth diagnosis list and a narrative variant, and issued a binary judgment of whether the diagnoses were supported by the narrative. This was repeated for all three narrative types across 90 cases. Cochran's Q test was used to assess global differences across conditions, and pairwise McNemar tests with Holm correction were used for post-hoc comparisons .

\subsubsection{Results.}  
Table~\ref{tab:diag_accept} summarizes results across the 90 paired cases. Cochran’s Q test found no significant differences in acceptability across conditions, and pairwise McNemar tests likewise show no meaningful differences between any pair of narrative variants.

\subsubsection{Interpretation.}  
GPT-5 judged all three variants as comparably adequate clinical evidence, further demonstrating that \textbf{Anonpsy preserves the core symptomatology, temporal evolution, and causal structure required for DSM-5-TR diagnostic reasoning}.

\begin{table}[t]
\centering
\small
\begin{tabular}{lcc}
\toprule
\textbf{Test} & \textbf{Statistic} & \textbf{$p$} \\
\midrule
Cochran's Q & 2.137 & 0.343 \\
\midrule
\multicolumn{3}{c}{\textbf{McNemar (Holm)}} \\
Anonpsy vs.\ Original & & 0.808 \\
Anonpsy vs.\ LLM-only & & 0.790 \\
LLM-only vs.\ Original & & 1.000 \\
\bottomrule
\end{tabular}
\caption{Diagnosis acceptability results across the three narrative variants.}
\label{tab:diag_accept}
\end{table}

\subsection{Semantic Similarity Analysis}
\label{sec:cosine}
For each case, we computed cosine similarity between the original narrative and each de-identified variant. Sentence-level embeddings were obtained using \texttt{all-mpnet-base-v2}, a Sentence-BERT model designed for semantic paraphrase identification \citep{reimers2019sentence}. Because SBERT explicitly maps paraphrases to nearby embedding vectors, cosine similarity provides an estimate of \emph{semantic} rather than merely lexical similarity.

Let $s_i^{\text{Anonpsy}}$ and $s_i^{\text{LLM}}$ represent the cosine similarity between the original narrative and each method for case $i$ and $d_i = s_i^{\text{Anonpsy}} - s_i^{\text{LLM}}$ denote the paired difference for case $i$. We tested whether Anonpsy yields significantly lower similarity to the original narrative, corresponding to the one-sided Wilcoxon signed-rank test with $H_0\!:\,\text{median}(d_i)=0$ and $H_1\!:\,\text{median}(d_i)<0$.

\subsubsection{Results.}  
Results are summarized in Table~\ref{tab:cosine-sim}. According to one-sided Wilcoxon signed-rank test, Anonpsy output showed significantly lower cosine similarity compared to LLM-only output.

\subsubsection{Interpretation.}
Anonpsy narratives are substantially more semantically distant from their originals
compared to the LLM-only narratives, with an extremely significant signed-rank result  
($p < 0.0001 $). This indicates that Anonpsy performs stronger semantic transformation, reducing the likelihood of re-identification due to similar narratives.

\begin{table}[!ht]
    \centering
    \small
    \begin{tabular}{lccc}
        \toprule
        \textbf{Method} & \textbf{Mean cos.\ sim.} & \textbf{Wilcoxon $p$} \\
        \midrule
        Anonpsy  & 0.718 & $<0.001$ \\
        LLM-only & 0.790 & -- \\
        \bottomrule
    \end{tabular}
    \caption{
    Semantic cosine similarity between original narratives and their de-identified
    counterparts.}
    \label{tab:cosine-sim}
\end{table}

\subsection{GPT-5 Re-identification Simulation}
\label{sec:gpt:reid}

The GPT-5 simulation mirrors the human re-identification protocol while treating GPT-5 as an informed insider. GPT-5 was first primed with the original narrative as system-level context and instructed to behave as someone who knows this patient. In the user turn, it received an A/B pair of de-identified narratives (Anonpsy vs.\ LLM-only in randomized order) and was asked to (i) select which version most strongly evoked the same patient and (ii) assign 1--5 re-identification risk scores to both versions using the same rubric as in the human study (Table~\ref{tab:rubric}). As in the human analysis, scores 1--2 were interpreted as "no or minimal" re-identification risk, and scores $\geq 3$ as progressively higher levels of recallable narrative identity. We analyzed paired GPT-5 risk scores with a two-sided Wilcoxon signed-rank test (Table~\ref{tab:reid-risk}).

\begin{table}[t]
    \small
    \begin{tabular}{lccc}
        \toprule
        \textbf{Metric} & \textbf{Anonpsy} & \textbf{LLM-only} & \textbf{ $p$} \\
        \midrule
        More recallable (\%) & 33.3\% & 66.7\% & -- \\
        Mean risk score      & 1.656  & 1.989  & 0.002 \\
        \bottomrule
    \end{tabular}
    \caption{
    GPT-5-based re-identification simulation: recallability and risk scores for Anonpsy and LLM-only narratives across 90 cases.
    }
    \label{tab:reid-risk}
\end{table}

\section{Component-Level Ablation Study}
\label{sec:ablation_details}

To identify contribution for each component, we conducted targeted ablation experiments isolating two core design elements of Anonpsy: (i) STEB-based symptom decomposition and (ii) retrograde perturbation ordering. All ablations were evaluated on the full set of 90 DSM-5-TR Clinical Cases using the identical GPT-5-based evaluation protocol described in Appendix~\ref{sec:gpt:diag:pred},~\ref{sec:gpt:diag:acc},~\ref{sec:cosine}, and~\ref{sec:gpt:reid}. No evaluation prompts, canonicalization rules, or statistical procedures were modified for ablation runs.

\subsection{Ablation Design}

\subsubsection{(a) STEB Decomposition Ablation (NoSTEB)}

In the full Anonpsy pipeline, each symptom node is decomposed into STEB components prior to perturbation. To isolate the contribution of this decomposition, we constructed a \textbf{NoSTEB} variant with the following modification:

\begin{itemize}
    \item During conversion ($\mathcal{E}$), instead of extracting structured STEB fields, we prompted the model to extract a single episode-level sentence per symptom node (if present).
    \item During perturbation ($\mathcal{P}$), this sentence was rewritten using the same retrograde ordering and demographic constraints as the full system, but without structured STEB fields.
    \item All temporal representations, typed relations, demographic perturbations, test-result substitutions, and graph-conditioned generation procedures were unchanged.
\end{itemize}

This isolates the structural contribution of STEB decomposition while preserving the remainder of the pipeline.

\subsubsection{(b) Retrograde Perturbation Ablation (NoRetro)}

In the full system, symptom contexts are perturbed in a temporally retrograde order (most recent to earliest), allowing later clinically salient episodes to anchor earlier rewrites. To test whether this ordering affects structural preservation, we constructed a \textbf{NoRetro} variant:

\begin{itemize}
    \item STEB decomposition was retained.
    \item Symptom contexts were perturbed in chronological (earliest-to-latest) order instead of retrograde order.
    \item All other components, including demographic perturbation, temporal normalization, graph structure, and generation, remained identical.
\end{itemize}

This isolates the effect of perturbation ordering on structural coherence.

\subsection{Evaluation Metrics}

All ablation variants were evaluated under the identical GPT-5-based protocol used for the full Anonpsy system (Appendix~\ref{sec:gpt:diag:pred},~\ref{sec:gpt:diag:acc},~\ref{sec:cosine},~\ref{sec:gpt:reid})

\subsection{Results}

\subsubsection{(a) STEB Decomposition Ablation (NoSTEB)}

Results are summarized in Table~\ref{tab:ablation_steb}.

\begin{table}[h]
\centering
\small
\begin{tabular}{lccc}
\toprule
\textbf{Metric} & \textbf{Full} & \textbf{NoSTEB} & \textbf{$p$} \\
\midrule
Mean Soft-F1 & 0.575 & 0.576 & 0.981 \\
Diag. Acceptability & 0.765 & 0.722 & 0.677 \\
Mean Cosine Similarity & 0.718 & 0.746 & $<0.001$ \\
Mean Risk Score & 1.656 & -- & $<0.001$ \\
\bottomrule
\end{tabular}
\caption{STEB ablation results (NoSTEB).}
\label{tab:ablation_steb}
\end{table}

Diagnostic predictability and acceptability were unchanged, whereas semantic similarity and re-identification risk increased significantly without STEB decomposition. This indicates that STEB primarily enhances privacy robustness while preserving clinical structure. 

\subsubsection{(b) Retrograde Perturbation Ablation (NoRetro)}

Results are summarized in Table~\ref{tab:ablation_retro}.

\begin{table}[h]
\centering
\small
\begin{tabular}{lccc}
\toprule
\textbf{Metric} & \textbf{Full} & \textbf{NoRetro} & \textbf{$p$} \\
\midrule
Mean Soft-F1 & 0.575 & 0.525 & 0.048 \\
Diag. Acceptability & 0.765 & 0.644 & 0.032 \\
Mean Cosine Similarity & 0.718 & 0.720 & 0.305 \\
Mean Risk Score & 1.656 & 2.070 & 0.748 \\
\bottomrule
\end{tabular}
\caption{Retrograde perturbation ablation results (NoRetro).}
\label{tab:ablation_retro}
\end{table}

Removing retrograde ordering significantly reduced diagnostic predictability and acceptability without materially affecting surface similarity. This indicates that retrograde perturbation primarily supports structural coherence rather than recallability reduction. 

\subsection{Complementary Roles of Components}

The two ablations reveal a functional dissociation:

\begin{itemize}
    \item \textbf{STEB decomposition} primarily enhances privacy robustness by reducing semantic overlap while preserving diagnostic structure.
    \item \textbf{Retrograde perturbation} primarily preserves clinical structure by stabilizing temporal and causal consistency during rewriting.
\end{itemize}

Together, these findings provide empirical justification for the architectural design of Anonpsy. The graph representation is not a superficial intermediate format; rather, its internal structure enables separable control over privacy robustness and clinical fidelity.

\section{Error Analysis}
\label{sec:app-err}

To better understand residual failure modes, we manually inspected output graphs and generated narratives that received relatively high re-identification risk scores in expert and automated (GPT-5) evaluations. This inspection revealed that the conversion stage occasionally adopts a conservative inclusion strategy, prioritizing clinically relevant entities even when their diagnostic certainty is uncertain.

One illustrative case involved anxiety about chronic Lyme disease. In the original narrative, the patient persistently believed he had Lyme disease despite repeated negative test results. Because the narrative repeatedly referenced Lyme disease using language typical of factual medical history (e.g., ``chronic illness'' and long-term disease coping), the semantic graph conversion appears to have encoded it as a past-history diagnosis node (\emph{``Chronic Lyme disease''}). As a result, the perturbation stage preserved this concept as part of the clinical structure, and the generated narrative continued to reference Lyme disease. Although demographic attributes and contextual life events were substantially altered, the persistence of this distinctive illness label allowed a recognizable narrative anchor to remain.

This example highlights a broader ambiguity in psychiatric narratives: subjective experiences, suspected conditions, and patient beliefs about medical illness are often described using the similar linguistic forms as confirmed medical diagnoses. For example, expressions such as ``feeling depressed,'' references to a feared illness, or patient-reported disease labels may linguistically resemble documented medical history even when they represent subjective interpretations or unverified conditions. When the conversion stage conservatively encodes such medically framed concepts as structured entities, structure-preserving perturbation may retain identity-bearing concepts that are not strictly required for diagnostic validity. Future work could address this limitation by explicitly distinguishing confirmed medical history from patient-believed illness targets, suspected conditions, or subjective symptom interpretations during graph construction, enabling selective perturbation when these concepts primarily function as narrative anchors rather than clinically necessary facts.

\end{document}